\documentclass[runningheads]{llncs}
\usepackage{comment}
\usepackage{graphicx}
\usepackage{subfigure}
\usepackage{cite}
\begin{document}
\title{Visible and NIR Image Fusion Algorithm Based on Information Complementarity}

\def\CICAISubNumber{429}  % Insert your submission number here
%\begin{comment}
\titlerunning{Fusion Algorithm Based on Information Complementarity}
\authorrunning{Zhuo Li et al.}
\author{Zhuo Li, Bo Li}
\institute{Beihang University, Beijing Key Laboratory of Digital Media}
%\end{comment}
%******************

\begin{comment}
\titlerunning{Abbreviated paper title}
\author{First Author\inst{1}\orcidID{0000-1111-2222-3333} \and
Second Author\inst{2,3}\orcidID{1111-2222-3333-4444} \and
Third Author\inst{3}\orcidID{2222--3333-4444-5555}}
%
\authorrunning{F. Author et al.}
\institute{Princeton University, Princeton NJ 08544, USA \and
Springer Heidelberg, Tiergartenstr. 17, 69121 Heidelberg, Germany
\email{lncs@springer.com}\\
\url{http://www.springer.com/gp/computer-science/lncs} \and
ABC Institute, Rupert-Karls-University Heidelberg, Heidelberg, Germany\\
\email{\{abc,lncs\}@uni-heidelberg.de}}
\end{comment}
%******************
\maketitle              % typeset the header of the contribution

\begin{abstract}
Visible and near-infrared(NIR) band sensors provide images that capture complementary spectral radiations from a scene. And the fusion of the visible and NIR image aims at utilizing their spectrum properties to enhance image quality. However, currently visible and NIR fusion algorithms cannot well take advantage of spectrum properties, as well as lack information complementarity, which results in color distortion and artifacts. Therefore, this paper designs a complementary fusion model from the level of physical signals. First, in order to distinguish between noise and useful information, we use two layers of the weight-guided filter and guided filter to obtain texture and edge layers, respectively. Second, to generate the initial visible-NIR complementarity weight map, the difference maps of visible and NIR are filtered by the extend-DoG filter. After that, the significant region of NIR night-time compensation guides the initial complementarity weight map by the arctanI function. Finally, the fusion images can be generated by the complementarity weight maps of visible and NIR images, respectively. The experimental results demonstrate that the proposed algorithm can not only well take advantage of the spectrum properties and the information complementarity, but also avoid color unnatural while maintaining naturalness, which outperforms the state-of-the-art.

\end{abstract}
\section{Introduction}

\begin{figure*}[!htp]
\centering
\subfigure[]{
\centering
\includegraphics[width=5.3cm,height=3.5cm]{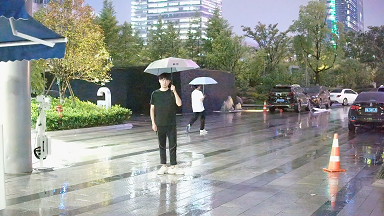}
\label{img1}
}%
\hspace{-0.3cm}
\subfigure[]{
\centering
\includegraphics[width=5.3cm,height=3.5cm]{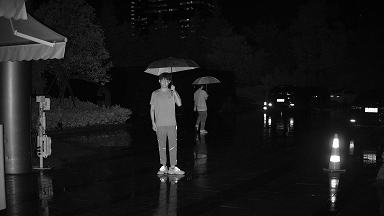}
\label{img2}
}%
\centering
\caption{Visible and NIR image captured by adding near-infrared light compensation }
\label{img3}
\end{figure*}

Recent studies have demonstrated various strategies for concurrently acquiring visible and Near-infrared (NIR) images, utilizing silicon-based sensors as imaging technology advances\cite{LI2017100, 1997Part}. Especially for low-light enhancement, utilizing two image sensors with specialized optical components, one of which takes visible spectra while the other captures near-infrared spectra by adding near-infrared light correction, as shown in Fig. \ref{img3}. The information of the two spectra can be combined by image fusion to generate high-quality images in various applications for night vision systems\cite{Zheng2012An,2019Adaptive,2020Low,2021Visible}.

For combining NIR information with visible images, numerous image fusion techniques have been developed, including traditional\cite{Sappa2016Wavelet, Vanmali2017Visible, Shutao2013Image, Kumar2015Image} and deep-learning methods \cite{LI2017100, 9628066}. The traditional methods focus on the intensity channel of the source images. The color information of visible spectra is retained by determining the chroma terms, finally used to reform the greyscale fused image into a color image.

In \cite{article}, bilateral and weighted least squares filters were employed by Sharma et al. to produce images that were upgraded with greater detail information and higher contrast. By taking into account the structural variations between visible and NIR images, Elliethy et al.\cite{8308623} method maintained the essential detail and edge information of a visible image. \cite{Son2017Near} proposed a mapping model that maintains local contrast in the NIR image while changing the NIR image values to correspond to the visible image corresponding pixels in the luminance plane. In \cite{Son2017Near2}, the relative difference of the local contrasts in the visible and NIR image was used to estimate a fusion map initially. In order to create the enhanced fusion image, the approach first extracted spatial features from the NIR image. Then the details were weighted in accordance with the fusion map and combined with the visible image. Additionally, Connah et al. proposed the image fusion technique based on the Spectral Edge(SpE)\cite{ConnahDF14} is a promising algorithm for combining images in the gradient domain. The kind of fusion method, however, requires a lot of computation and iteration.

Some data-driven methods were developed based on the deep learning theory\cite{Multi-spectralJ,9628066,2021Deep}. For the fusion of visible and infrared images, Li et al.\cite{Multi-spectralJ} suggested a deep learning system with an encoder, fusion layer, and decoder model. And the Dark Vision Net (DVN)\cite{DBLP}, a near-infrared visible light fusion method based on Deep Structure and Deep Inconsistency Prior (DIP) was proposed. Deep Structure is used to extract structural information from multi-scale features of extracted depth, avoiding introducing a lot of noise directly from the original input. Then based on this structural information, DIP uses structural inconsistency to guide the fusion of visible-NIR. The above method based on deep learning has made remarkable achievements, but its defect is that it requires a large number of training samples and computing power. The quality of image fusion depends on the training label material, model capacity, loss function selection, super parameter selection, and many other aspects.

Overall, the above-mentioned schemes require a color compensation process to generate a natural color-looking image. For many applications, it is desirable to retain the characteristic colors of the scene. However, modifying the chroma terms often leads to undesirable false-color image renderings\cite{5650293}. Therefore, we think more attention should be paid to the analysis of complementary at the level of image data, which is a display of physical signals. This paper designs a complementary fusion model from the level of physical signals. First, in order to distinguish between noise and useful information, we use two-scale image filters to obtain texture and edge layers, by the weight-guided filter and guided filter in proper order. Second, to generate the initial complementarity weight map, the difference maps of visible and NIR are filtered by the extend-DoG filter. After that, the significant region of NIR guides the initial complementarity weight map by the arctanI function. Finally, the fusion images can be generated by the complementarity weight maps of visible and NIR images, respectively. The experimental results demonstrate that the proposed algorithm can not only well take advantage of the properties of different spectra and the information complementarity, but also avoid color distortion while preserving chromaticity information of visible images, which outperforms the state-of-the-art.

The rest of this paper is organized as follows. The complementary theory and related work above the visible and near-infrared (NIR) are both covered in Section II. The specifics of the proposed fusion algorithm are described in Section III. Section IV assesses both the objective and subjective performance of the suggested algorithm. Finally, Section V concludes the paper.

\section{Related Work}

As we have discussed in\cite{2020Spectrum}, each spectral band has different spectrum properties, providing different kinds of physical information. The similarity and difference between visible and NIR images have been addressed in\cite{2017CORRELATION, 8918077}. The authors computed the correlation between visible and NIR gradients, and use the gradients of the visible image in reconstructing NIR only where the gradients are highly correlated. Meanwhile, many researchers often assumed that the high-frequency information of the visible-NIR channels is strongly correlated in\cite{Brown2011Multi, Salamati2010Material}.

Moreover, according to the theory of spectral reflection, the difference between visible and NIR spectra is the reflection-dependent molecular clusters\cite{Salamati2010Material, Zheng2012An}. The visible spectral reflection is based on the structural system in molecules called chromophores, therefore the visible images are colorful and suitable for human visual perception. However, the NIR spectral reflects the composition and molecular structure information of most types of organic compounds, namely NIR radiation is material dependent and with no color information.

It can be found that there are significant similarities between visible and near-infrared information that can be transformed, however, there are also some that cannot be obtained through transformation. In this paper, the information of one spectral cannot be transformed from the other one is called complementarity information. We think the color information of visible bands could not be replaced by NIR\cite{Fredembach2008Colouring}, therefore, the R, G, and B three bands of visible are combined with NIR band, respectively. Meanwhile, in low-light conditions utilizing two image sensors to capture visible spectra and near-infrared spectra by adding near-infrared light compensation, the NIR image contains a high signal-to-noise ratio, which can be supplemented for the visible image.

\begin{figure}
\includegraphics[width=\textwidth]{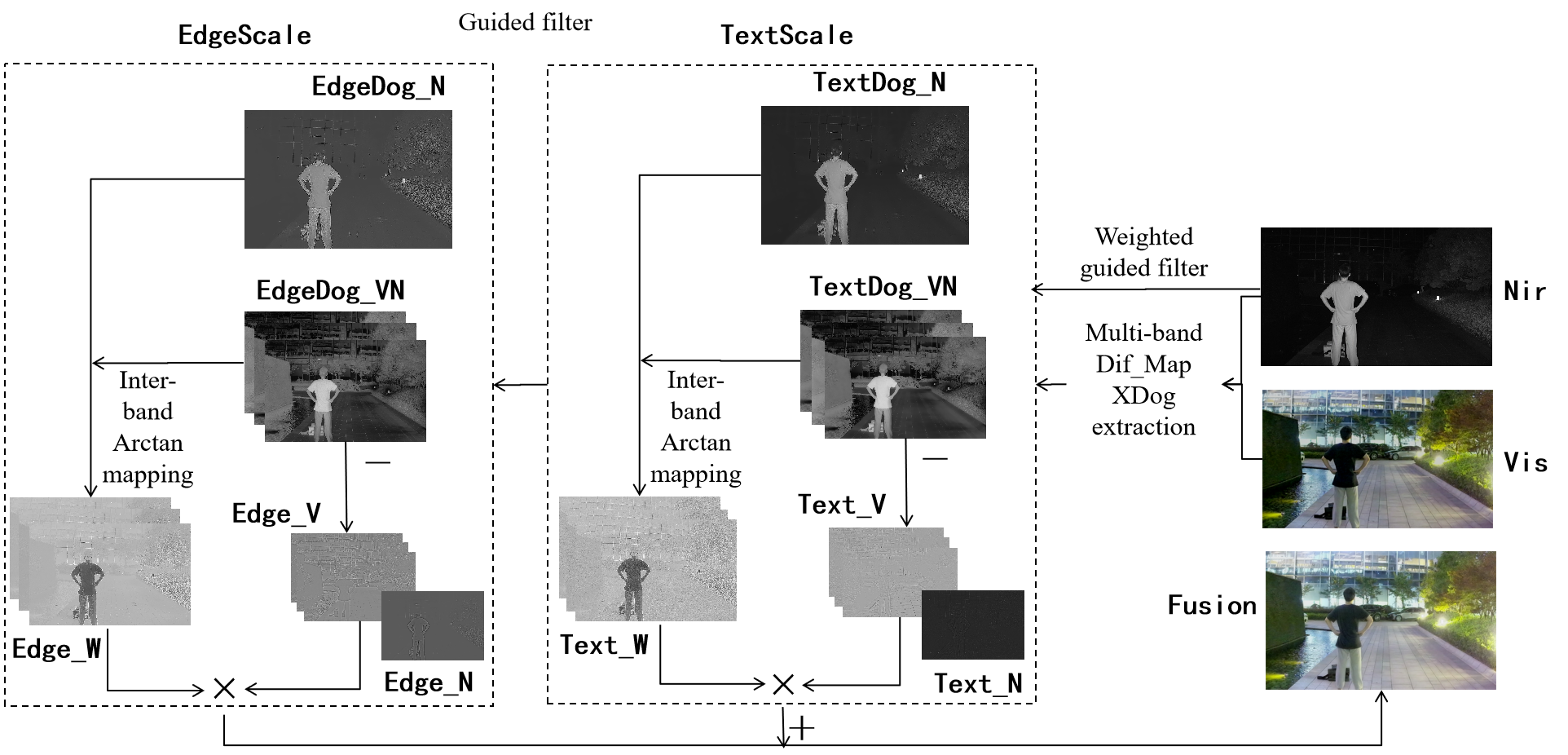}
\caption{The overall structure of the proposed fusion algorithm.} \label{fig0}
\end{figure}

\section{The proposed algorithm}
\subsection{Two-scale guided image decomposition}

The proposed fusion algorithm is shown in Fig. \ref{fig0}, in order to better distinguish between noise and useful information, as well as maximize the utilization of high-frequency information, we use two-scale image filters to obtain texture layers and edge layers, by weighted guided and guided filter respectively. As shown in Fig. \ref{img7},  we can see that the texture layer has more noise than the edge layer, which is more clear.

Considering the dark and uneven illumination conditions at night, the high-frequency noise of visible images in the texture layer is high. In order to distinguish noise and texture information in the first layer, we adopt the weight-guided filter(WGIF) proposed in \cite{2015Weighted}, which can preserve sharp edges like the global filters, similar to the guided filter(GIF) in \cite{Shutao2013Image}. Besides that, the WGIF also avoids gradient reversal, therefore the halo artifacts can be avoided by the WGIF. After the first layer of filtering decomposition, the second layer filtered by the GIF will contain clear edge information. To some extent, the first layer of filtering can reduce the gradient of large-scale edges, which will also reduce the halo phenomenon of the second layer of filtering. The two-scale filters process is as follows:

\begin{figure*}[!htp]
\centering
\subfigure[]{
\centering
\includegraphics[width=5.5cm,height=4cm]{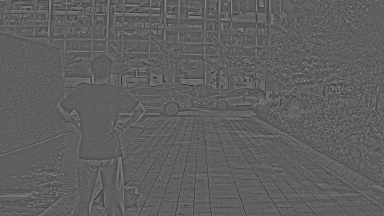}
\label{img5}
}%
\hspace{-0.3cm}
\subfigure[]{
\centering
\includegraphics[width=5.5cm,height=4cm]{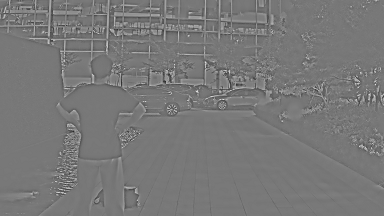}
\label{img6}
}%
\centering
\caption{The texture and edge layers of the Red channel of the visible image.}
\label{img7}
\end{figure*}

\begin{equation}\label{1}
  R_p^1=WGuidF_{\tau,\varepsilon}(R_p^0,R_p^0),\quad R_p^2=GuidF_{\omega,\theta}(R_p^1,R_p^1)
\end{equation}

where $p$ represents the R, G, B channels of visible and NIR channel index, $p\in{\{r,g,b,n\}}$; $R_p^1$ are the first base layers calculated by the weight-guided filter; $R_p^2$ are the second base layers from the guided filter. The reason for using the weight-guided filter in the first layer is that it can better distinguish noise and preserve the gradients of edges. The $\tau$, $\varepsilon$, $\omega$, and $\sigma$ are the parameters of the filter functions.

\begin{equation}\label{2}
    D_p^{i+1}={R_p^i}-{R_p^{i+1}}
\end{equation}

where $D_p^{i+1}$ means the detail and edge layers of the image, $p$ is the channel index, represents R, G, and B of visible and NIR channels, and $i$ in this paper is $i=0,1$. That is to say, $D_p^{1}$ means the detail layer.

\subsection{Inter-band Information Complementarity Map Estimatation}

The relationship between the visible and NIR spectral bands has been discussed in Section II, particularly in the context of the reflectivity of vegetation and other natural scene features. It can be clearly seen that the original near-infrared is significantly different from all three original visible bands.

According to \cite{2004Understanding}, the difference maps of visible and NIR images can include information from two modalities. The smaller the difference, the closer the indication is. The larger the difference, the more difficult. And it is also applied to the extraction of the DOG filter of difference maps. Therefore, We use the feature of difference maps to esitmate the initial visible-NIR complementarity map, as well as add the significant region of near-infrared night-time compensation to guide the complementary information.

Additionally, the fundamental layer's potential texture and edge guide map are obtained using the extend-DoG((XDog))\cite{WINNEMOLLER2012740, 40534} method, the main goal of which is to expand the edge information. The edge line width of the generated edge graph is usually only 1¡¢2 pixels, however, the general edges of images should not be so thin. The color weight map is obtained by using extend-DoG and the complementary information is obtained from the near-infrared light supplement area at night. The extend-DoG filter is applied to estimate the difference maps as the initial complementarity weight maps as follows:

\begin{equation}\label{3}
  RN_c^i=Normal(R_c^i-R_n^i),\quad
\end{equation}
\begin{equation}\label{4}
  dog_c^i=XDoG(RN_c^i),\quad
  dog_n^i=XDoG(R_n^i),\quad
\end{equation}

where $Normal()$ means the normalization processing of the base layers of the visible and NIR images processed by two-scale filters, $i=0,1$ represents the texture and edge layer, $c$ is the R, G, and B of visible channels, $n$ is the NIR channel.

\begin{figure*}[!htp]
\centering
\subfigure[]{
\centering
\includegraphics[width=4cm,height=3cm]{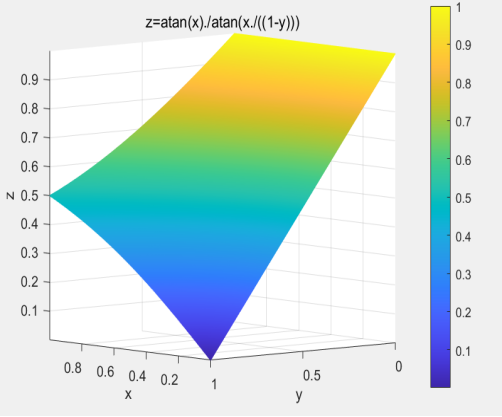}
\label{img141}
}%
\hspace{-0.4cm}
\subfigure[]{
\centering
\includegraphics[width=4cm,height=3cm]{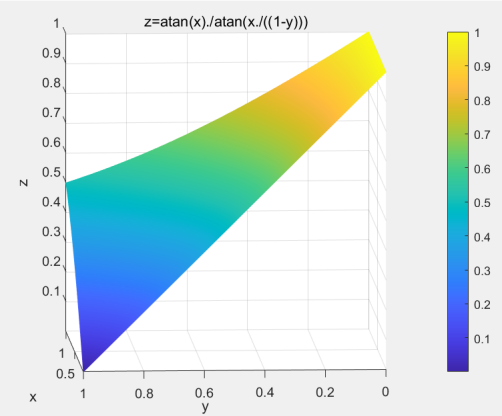}
\label{img142}
}%
\hspace{-0.4cm}
\subfigure[]{
\centering
\includegraphics[width=4cm,height=3cm]{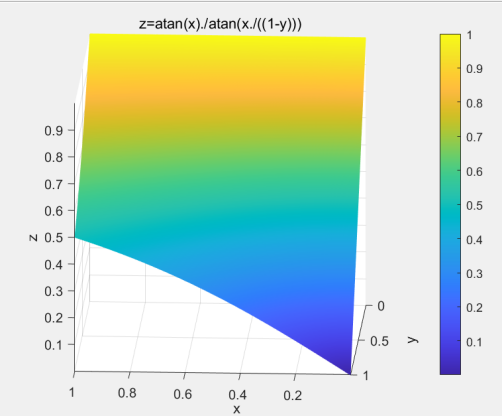}
\label{img143}
}%
\centering
\caption{The arctanI function used to fit the information complementary}
\label{img10}
\end{figure*}

\subsection{Information Complementary Weight Model}

Due to the low illumination data, when the brightness value in the image area is relatively low, the noise will be higher and the dog value will also be higher; So only when the brightness value of the image area is relatively high and the dog value is large, will useful information be obtained. So, we believe that the XDog value of an image, combined with its pixel value size, reflects whether it is a texture with a high signal-to-noise ratio or information with a low signal-to-noise ratio. As a preliminary estimate of fusion weights.

Considering that at night, the signal-to-noise ratio of near-infrared images that rely on supplementary lighting will be relatively high in the supplementary area, but the signal-to-noise ratio of areas that do not receive supplementary lighting is still relatively low. So, we use the extend-DoG filter and the value of near-infrared as a variable factor for adjustment.

\begin{theorem}
When the signal-to-noise ratio of near-infrared is high, the weight value of R, G, and B of visible channels needs to be changed within a lower range based on the signal-to-noise ratio of visible light.
\end{theorem}
\begin{theorem}
When the signal-to-noise ratio of near-infrared is low, the weight value of R, G, and B of visible channels needs to be changed within a higher range based on the signal-to-noise ratio of visible channels.
\end{theorem}

Inspired by\cite{WINNEMOLLER2012740}, the arctanI function is used for adjustment to fit the information complementary between visible and NIR images as follows:

\begin{equation}\label{5}
 fuwt_c^i=arctanI(dog_c^i,dog_n^i)
\end{equation}

where $arctanI$ is the function used to control the complementary based on the signal-to-noise of NIR, it is defined as follows:

\begin{equation}\label{6}
arctanI(x,y)=\frac{atan(x)}{atan(\frac{x}{1-y})+\alpha}
\end{equation}
As shown in the Fig. \ref{img10}, when the $y$ value of NIR $dog_n^i$ is a range of 0.5 approaches 1, the $x$ value of visible bands $dog_c^i$ varies between 0 and 0.5; When the $y$ value of $dog_n^i$ is less than 0.5 and close to 0, the $x$ value of visible bands $dog_c^i$ changes between 0.5 and 1.

After the weights $fuwt_c^i$ of visible channels have been generated, the fused detail and edge layers can be implemented from each visible channel and NIR according to their own weights as follows:
\begin{equation}\label{7}
fuD_{c}^{i+1}=D_{c}^{i+1}.*fuwt_{c}^{i+1}+D_{n}^{i+1}.*(1-fuwt_{c}^{i+1}))
\end{equation}

Finally, combining the fused detail and edge layers and the correspondent base layer of RGB, the fusion $ F_{c}$ is defined as:

\begin{equation}\label{8}
  F_{c}=R_{c}^{2}+fuD_{c}^{2}+fuD_{c}^{1}
\end{equation}

\begin{figure*}[!htp]
\centering
\subfigure[]{
\centering
\includegraphics[width=3cm,height=2cm]{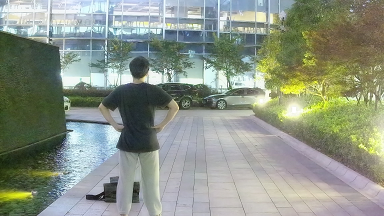}
\label{img151}
}%
\hspace{-0.4cm}
\subfigure[]{
\centering
\includegraphics[width=3cm,height=2cm]{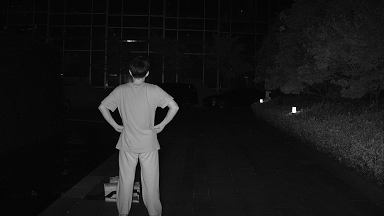}
\label{img152}
}%
\hspace{-0.4cm}
\subfigure[]{
\centering
\includegraphics[width=3cm,height=2cm]{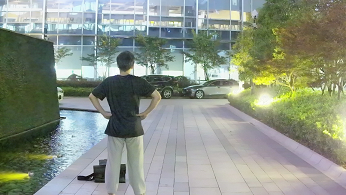}
\label{img153}
}%
\hspace{-0.4cm}
\subfigure[]{
\centering
\includegraphics[width=3cm,height=2cm]{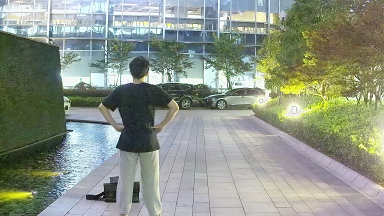}
\label{img154}
}
\vfill   %Õâ¸ö»Ø³µ¼üºÜÖØÒª \quadÒ²¿ÉÒÔ
\subfigure[]{
\centering
\includegraphics[width=3cm,height=2cm]{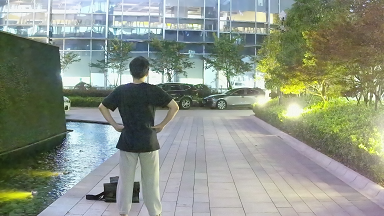}
\label{img155}
}%
\hspace{-0.4cm}
\subfigure[]{
\centering
\includegraphics[width=3cm,height=2cm]{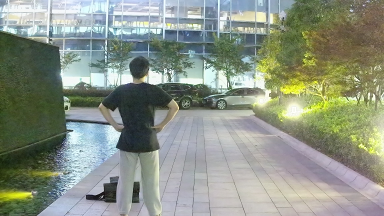}
\label{img156}
}%
\hspace{-0.4cm}
\subfigure[]{
\centering
\includegraphics[width=3cm,height=2cm]{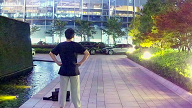}
\label{img157}
}%
\hspace{-0.4cm}
\subfigure[]{
\centering
\includegraphics[width=3cm,height=2cm]{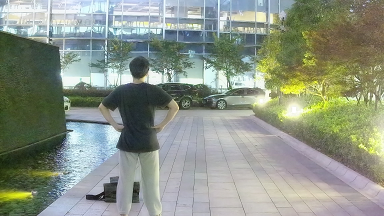}
\label{img158}
}%
\centering
\caption{Qualitative comparison of the outdoor image (a) Visible image. (b) Near-infrared image. The fusion images are obtained by (c) BF\cite{Kumar2015Image} (d) CP\cite{2019Adaptive} (e) JSM\cite{Yan2013Cross} (f) Lap\cite{Vanmali2017Visible}  (g) SE\cite{ConnahDF14} (h) our algorithm.}
\label{img15}
\end{figure*}

\section{Experimental results}
In this paper, by taking advantage of the spectrum properties and the information complementarity in fusion processing, the proposed fusion algorithm can not only improve image quality but also avoid color distortion while maintaining the natural color look. To verify the advantages of our fusion algorithm, We compare the proposed method with traditional state-of-the-art fusion methods including Joint Scale Map Restoration fusion(JSM)\cite{Yan2013Cross}, Bilateral Filter fusion(BF) \cite{Kumar2015Image}, Color Preserve fusion(CP)\cite{2019Adaptive}, Laplacian Pyramid fusion(Lap)\cite{Vanmali2017Visible} and Spectral Edge fusion(SE) \cite{ConnahDF14}. And the group of test images is pairs of visible and NIR images captured by Hikvision Black Light Camera, which can simultaneously capture visible and NIR images with two CCDs through the same optical path. And we collected data on different conditions of lighting and noise, indoors and outdoors respectively.

\subsection{Objective comparison}
In our experiments, in order to value the authenticity and naturalness color of fusion images, we employ two metrics. To confirm the spectrum properties preservation as well as the chromaticity information of the visible image, the no-reference spectrum distortion index (SDI) is adopted\cite{2020Spectrum}, it is used to determine whether the correlation between visible and near-infrared spectra has changed.

\begin{figure*}[!htp]
\centering
\subfigure[]{
\centering
\includegraphics[width=3cm,height=2cm]{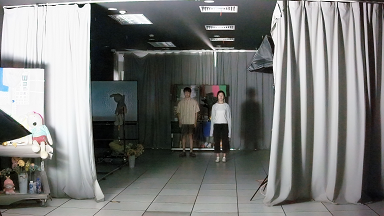}
\label{img161}
}%
\hspace{-0.4cm}
\subfigure[]{
\centering
\includegraphics[width=3cm,height=2cm]{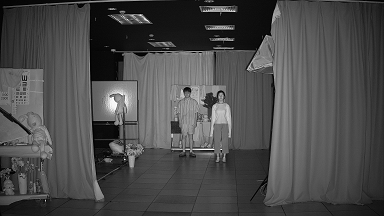}
\label{img162}
}%
\hspace{-0.4cm}
\subfigure[]{
\centering
\includegraphics[width=3cm,height=2cm]{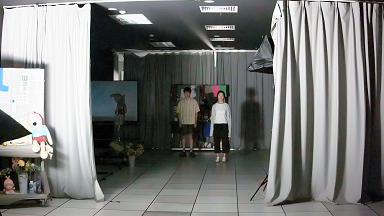}
\label{img163}
}%
\hspace{-0.4cm}
\subfigure[]{
\centering
\includegraphics[width=3cm,height=2cm]{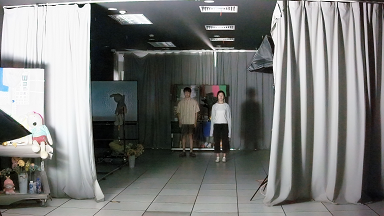}
\label{img164}
}
\vfill   %Õâ¸ö»Ø³µ¼üºÜÖØÒª \quadÒ²¿ÉÒÔ
\subfigure[]{
\centering
\includegraphics[width=3cm,height=2cm]{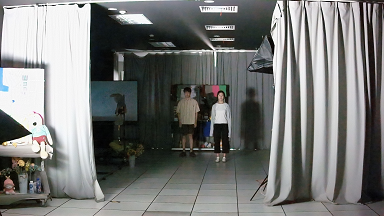}
\label{img165}
}%
\hspace{-0.4cm}
\subfigure[]{
\centering
\includegraphics[width=3cm,height=2cm]{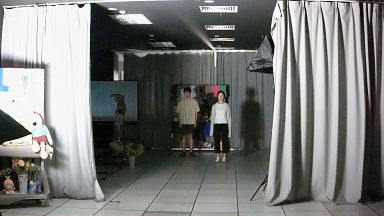}
\label{img166}
}%
\hspace{-0.4cm}
\subfigure[]{
\centering
\includegraphics[width=3cm,height=2cm]{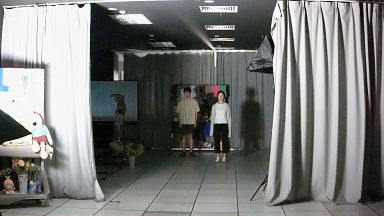}
\label{img167}
}%
\hspace{-0.4cm}
\subfigure[]{
\centering
\includegraphics[width=3cm,height=2cm]{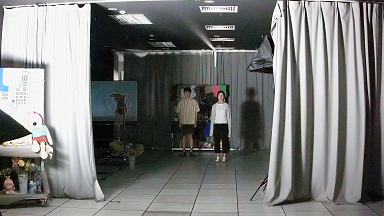}
\label{img168}
}%
\centering
\caption{Qualitative comparison of the indoor image. (a) Visible image. (b) Near-infrared image. The fusion images are obtained by (c) BF\cite{Kumar2015Image} (d) CP\cite{2019Adaptive} (e) JSM\cite{Yan2013Cross} (f) Lap\cite{Vanmali2017Visible}  (g) SE\cite{ConnahDF14} (h) our algorithm.}

\label{img16}
\end{figure*}

Besides that, we also adopt a simple metric of color distance(CD)\cite{2016Color}. This metric is a combination of both weighted Euclidean distance functions, where the weight factors depend on how significant the "red" component of the color is. This formula has results that are more stable algorithms, the selection of the closest color is subjective. And the smaller both values of the two metrics are, the less the correlations between R, G, and B of visible and near-infrared bands have changed. And the CD is defined as follows:

\begin{equation}\label{9}
 r=\frac{C_{c,r}+C_{f,r}}{2},\quad
\end{equation}

\begin{equation}\label{10}
 R^{'}=C_{c,r}-C_{f,r},
 G^{'}=C_{c,g}-C_{f,g},
 B^{'}=C_{c,b}-C_{f,b},
\end{equation}

\begin{equation}\label{11}
 CD=\sqrt{(2+\frac{r}{256})\times R^{'2}+4\times G^{'2}+(2+\frac{255-r}{256})\times B^{'2}}
\end{equation}

Meanwhile, to verify that the fused information comes from the original image, rather than artificially adding false information, we used SSIM\cite{1292216} to measure the structural similarity between the source image and the fusion image. In addition, the peak signal-to-noise ratio (PSNR) is used to quantify the quality of the fusion image, as well as the visual information fidelity (VIF) \cite{LI2017100} is used to measure information fidelity by calculating the distortion between the source image and the fusion image.

Table I demonstrates the comparisons of test images with corresponding algorithms, including 5 metrics. The greater value of the VIF, SSIM, and PSNR, the better the performance. As expected, the SE and Lap algorithms have high scores of SDI and CD, respectively. As shown in Fig. \ref{img15}(g) the color is not real, as well as the other values of metrics are not well. These algorithms, in general, aim on maintaining more details, nonetheless, the colors are oversaturated and unrealistic.

Although the SE and JSM algorithms have lower values of SDI and CD, the proposed algorithm can have the best PSNR value and the second VIF value. That is to say, the proposed algorithm can not have a color appearance suitable for the human eye, but also ensure the clarity of the fusion image, avoiding high-frequency noise. Meanwhile, the highest value of SSIM of the proposed algorithm indicates the fused information has the greatest correlation with the source information.

With a comprehensive examination of indicator data, our algorithm can take advantage of the correlation and complementarity between visible and NIR spectra, resulting in a natural color look. While the color appearance of the SE and JSM algorithms is closest to the original viewable image, the edges and dark areas are not obvious.

\begin{table*}[!htp]
\centering
\caption{TABLE I QUALITATIVE COMPARISION}
\begin{center}
\begin{tabular}{|l|l|l|l|l|l|l|}
\hline
\textbf{Metrics}&\multicolumn{6}{|c|}{\textbf{ COMPARISION ALGORITHM}} \\
\cline{2-7}
\textbf{}&\textbf{\textit{JSM\cite{Yan2013Cross}}}&\textbf{\textit{SE\cite{ConnahDF14}}}&\textbf{\textit{Lap\cite{Vanmali2017Visible}}}&\textbf{\textit{BF\cite{Kumar2015Image}}}&\textbf{\textit{CP\cite{2019Adaptive}}}&\textbf{\textit{Ours}} \\
\hline
CD& 0.98&5.00 &5.72 &1.11 &0.76 &0.73  \\
\hline
SDI& 0.0021&0.0777 &0.0421 &0.0321 &0.0020 &0.0018  \\
\hline
VIF& 0.41&0.38 &0.36 &0.39 &0.57 & 0.46 \\
\hline
SSIM& 0.48&0.38 &0.41 &0.56 &0.63 &0.65  \\
\hline
PSNR& 39.83&37.46 &27.07 &37.46 &37.88 &40.93 \\
\hline
\end{tabular}
\label{tab1}
\end{center}
\end{table*}

\subsection{Subject comparison}

As shown in Fig. \ref{img15} and Fig. \ref{img16}, the fusion results with the SE, and the Lap algorithms all have color distortion, such as the clothes of people, and the areas of trees. Although the details in some dark areas can be seen with their algorithms, the color is unrealistic, such as the ground steps areas in Fig. \ref{img15}. These algorithms, in general, emphasize maintaining more details, however, the colors are over-saturated and artificial, particularly in places with large variations between visible and NIR spectra. Furthermore, the noise on the black garments of the guy in the scene has been reduced in our fusion results. This demonstrates that our algorithm is capable of distinguishing useful information from meaningless noise input.

More importantly, due to overexposure, the details of the visual acuity chart in the visible image are invisible, which can be seen in the NIR image, as shown in Fig. \ref{img16}.
However, in the fusion images, our visual acuity chart features are the most evident, whereas the others are either indistinct or not visible.
This comparison demonstrates that our algorithm can profit from the information complementarity between visible and NIR images, as well as distinguish valuable information from noise. Overall, our technology generates high-quality images with natural-looking color appearances.

\section{Conlusion}
In this paper, the fusion model based on the level of physical signals is proposed, with information complementarity between visible and NIR images. First, to distinguish between noise and useful texture and edge information, we use two layers of the weight-guided filter and guided filter to obtain texture and edge layers, respectively. Second, the extend-DoG filter is applied to estimate the difference maps as the Visible and NIR complementarity map. After that, to obtain the signal complementary weight, the significant region of NIR night-time compensation guides the complementarity map by the arctanI function we defined. Finally, the fusion images can be generated by the weights of three bands of visible and NIR images, respectively. Based on the physical optics imaging theory, the proposed algorithm analyzes the complementarity of physical information reflected in different light bands and designs a reasonable fusion model. Experiment findings show that our proposed fusion method outperforms state-of-the-art algorithms in both subjective and objective measures.

% References will then be sorted and formatted correctly.
%
\bibliographystyle{splncs04}

\bibliography{bibref}

\end{document}